\documentclass{article}

\usepackage[final]{timeseries_workshop}
\usepackage{graphicx}
\usepackage{amsmath}

\usepackage[utf8]{inputenc} 
\usepackage[T1]{fontenc}    
\usepackage{hyperref}       
\usepackage{url}            
\usepackage{booktabs}       
\usepackage{amsfonts}       
\usepackage{nicefrac}       
\usepackage{microtype}      
\usepackage{xcolor}         

\title{Zero Shot Time Series Forecasting Using Kolmogorov Arnold Networks}

\author{%
  Abhiroop Bhattacharya\\
  Department of Electrical Engineering\\
  École de Technologie Supérieure, Montreal, Canada.\\
  \texttt{abhiroop.bhattacharya.1@ens.etsmtl.ca} \\
  \And
  Nandinee Haq \\
  Hitachi Energy Research\\
  Montreal, Canada. \\
  \texttt{nandinee.haq@hitachienergy.com} \\
}

\begin{document}

\maketitle

\begin{abstract}
Accurate energy price forecasting is crucial for participants in day-ahead energy markets, as it significantly influences their decision-making processes. While machine learning-based approaches have shown promise in enhancing these forecasts, they often remain confined to the specific markets on which they are trained, thereby limiting their adaptability to new or unseen markets. In this paper, we introduce a cross-domain adaptation model designed to forecast energy prices by learning market-invariant representations across different markets during the training phase. We propose a doubly residual N-BEATS network with Kolmogorov Arnold networks at its core for time series forecasting. These networks, grounded in the Kolmogorov-Arnold representation theorem, offer a powerful way to approximate multivariate continuous functions. The cross domain adaptation model was generated with an adversarial framework. The model's effectiveness was tested in predicting day-ahead electricity prices in a zero shot fashion. In comparison with baseline models, our proposed framework shows promising results. By leveraging the Kolmogorov-Arnold networks, our model can potentially enhance its ability to capture complex patterns in energy price data, thus improving forecast accuracy across diverse market conditions. This addition not only enriches the model's representational capacity but also contributes to a more robust and flexible forecasting tool adaptable to various energy markets. 
\end{abstract}

\section{Introduction}

The increasing competitiveness of electricity markets has driven significant advancements in electricity price forecasting. Accuracy of the forecasts drive the bids for buying and selling electricity in the day-ahead market and hence reliable price forecasts are essential for market participants such as suppliers and traders. In markets where data is scarce or training could be costly, domain adaptation based machine learning techniques could offer solutions to generate forecasts for electricity prices in a zero-shot fashion without the requirement of training the model on the target market. While domain adaptation has seen successful applications in the field of computer vision, applying these methods to time series forecasting requires considerations to the temporal dynamics and local patterns inherent to time series \cite{li2023time}.

In the past years, several different time series specific models have emerged that focuses on learning the temporal patterns for forecasting. The Neural Basis Expansion Analysis model, N-BEATS \cite{oreshkin2019n} is one such model that has shown superior performance for domain-specific forecasting tasks. N-BEATS model architecture comprises of two main Multi-Layer Perceptron (MLP) based components: the backcast stack which processes historical data, and the forecast stack which predicts future values. In this paper, we propose an adversarial domain adaptation based framework for zero shot forecasting of electricity prices with an architecture based on N-BEATS model. Inspired by the Kolmogorov-Arnold representation theorem \cite{liu2024kan}, we propose integrating Kolmogorov-Arnold Networks (KANs) within the doubly residual architecture of the N-BEATS model for generalized feature extraction. KANs have emerged as a promising alternative to MLPs, which, unlike MLPs, utilize learnable activation functions on the edges by replacing the linear weights with univariate functions parametrized as splines. This architecture enables KANs to outperform MLPs in terms of accuracy and interpretability, achieving better results with fewer parameters and providing more intuitive visualizations. Our proposed model, built by integrating KANs with an N-BEATS-like architecture and trained with adversarial technique using a gradient reversal layer \cite{ganin2016domain}, ensures that the initial stack captures generalizable features useful for extracting domain-invariant representations, while deeper stacks focus on domain-specific features.

The key contributions of this paper are two-fold. This is the first work that leverages a combination of Kolmogorov Arnold networks and doubly residual connection networks like N-BEATS for time series forecasting. Building on this, we further propose an adversarial domain adaptation based framework for zero shot forecasting of energy prices by creating a generalized representation.

\section{Model Architecture}

\begin{figure}
    \centering
    \includegraphics[width=0.8\linewidth]{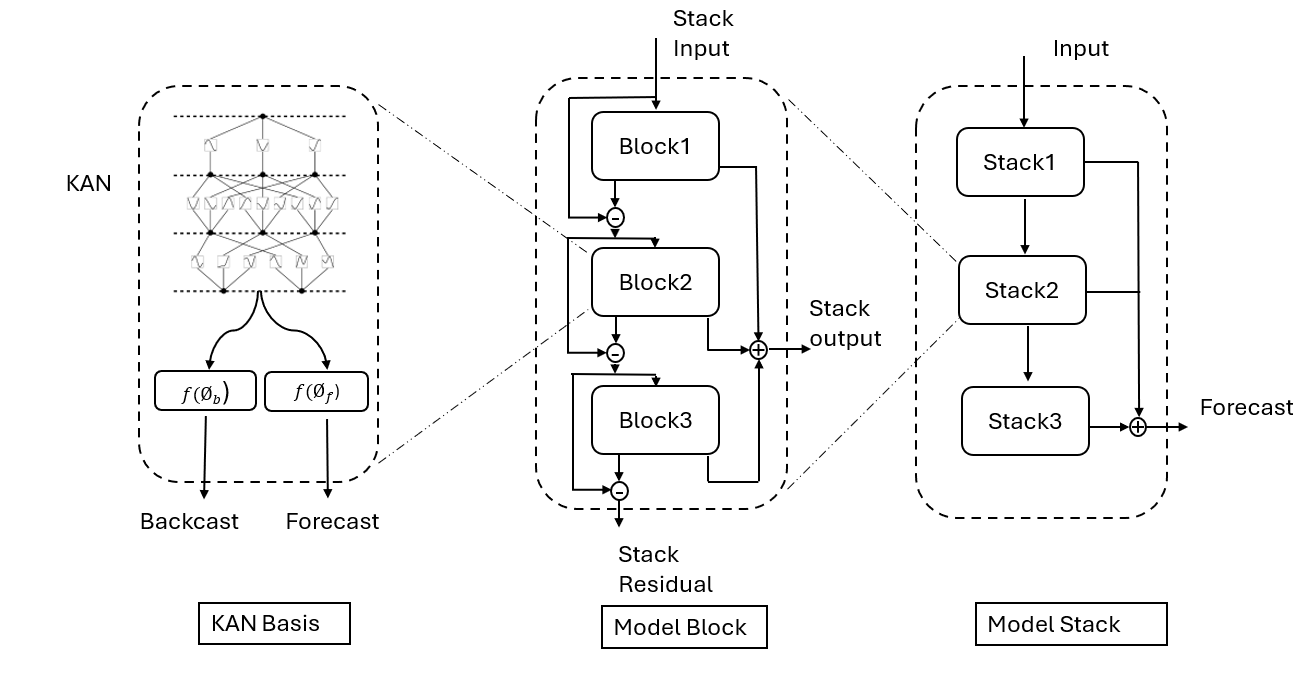}
    \caption{ Line Schematic showing the model architecture consisting of KAN layers stacked together using residual connections inspired by the N-BEATS architecture \cite{oreshkin2019n}}
    \label{fig:arch1}
\end{figure}

In this work,  we use a deep stack of Kolmogorov Arnold networks (KANs) with doubly residual connections. The network decomposes the time series into local projections by using univariate function parameters along the edge of the network. The Kolmogorov-Arnold representation theorem states that any multivariate continuous function can be decomposed into a finite sum of compositions of univariate functions. This allows KAN to model complex interactions in high-dimensional data through compositions of simpler univariate functions. KAN applies adaptive, learnable activation functions on the edges between nodes. These functions are parameterized as B-spline curves, which adjust dynamically during training to better capture the underlying data patterns.

Since the univariate functions are piecewise polynomials with specific degrees and global smoothness, they exhibit excellent approximation behavior relative to their degrees of freedom.  The doubly residual principle, inspired by the N-BEATS architecture \cite{oreshkin2019n}, is used between stacks. The time series is sequentially decomposed by  subtracting the predicted backcast $\hat{y}^{b}_{i,j}$ from the original series to obtain the next series $y^{b}_{i,j+1}$. The output of each forecast $\widehat{y}$ is obtained through hierarchical aggregation of each block's forecast and the last backcast derived by a residual sequence of blocks which serves as an input to the next stack. Fig. \ref{fig:arch1} shows the proposed model architecture where the model is composed of three sequential stacks to generate the overall forecasts. Each stack has three sequential blocks of neural networks, and each block consists of KAN layers that generate the backcast and forecast estimates, which are then fed onto the next block. 

This model architecture is used as the backbone for creating a generalized representation using a domain adaptation approach. We take day ahead prices from two established markets with significant historical data to generate the domain generalized model, and learn the generalized representation by using the supervised forecasting error on the primary market and a notion of feature distance between the primary and secondary market prices. This setup enables the model to learn the domain invariant features, which if given to a classifier, the classifier should not be able to predict which domain or market the features are originating from. In addition to the domain invariant features, we use the supervised training approach to learn domain or market specific features. To implement the adversarial training between the forecasting model and the domain classifier, we use a gradient reversal layer proposed by Ganin \textit{et.al.} \cite{ganin2016domain}.

\section{Dataset}
We train and evaluate our model’s forecasting capabilities using day ahead electricity prices from major power markets.  
Day ahead hourly electricity prices from the Nord Pool electricity market (NP), which corresponds to the Nordic countries exchange was taken as the target or unseen market for our experiments. The test period was from 1st January, 2018 to 24th December, 2018. One full year of test data was used to capture errors across all seasons.  
 Hourly electricity price data from three different markets were considered when training the domain-generalized models. The first train dataset is from the Pennsylvania-New Jersey-Maryland (PJM) market in the United States, which contains data from 1st January, 2013 to 24th December, 2018. The remaining two market prices are obtained from the integrated European Power Exchange (EPEX). The Belgium (BE) and French (FR) market data spans from 9th January, 2011 to 31st December, 2016.

\section{Results}

\begin{table}
  \caption{Comparison of zero shot performance for the Nord Pool Market.}
  \label{tab:zero_shot}
  \centering
  \begin{tabular}{rrrrrrrrr}
    \toprule
& \multicolumn{3}{c}{MAE} & & \multicolumn{3}{c}{SMAPE} \\
\midrule
Primary & KAN & N-BEATS & Proposed && KAN & N-BEATS & Proposed\\ \midrule
FR & 3.0649 & 2.7416 & \textbf{$2.5056 \pm 0.10$} && 0.1069 &	0.0932 & \textbf{$0.0854 \pm 0.003$}\\
PJM & 3.1942 & 3.2545 & \textbf{$2.5697\pm 0.12$} && 0.1110 &	0.1044 & \textbf{$0.0862 \pm 0.002$}\\
BE &3.1409& 2.5904& \textbf{$2.5144 \pm 0.09$} &&	0.1094 & 0.0869 & \textbf{$0.0857 \pm 0.004$}\\
    \bottomrule
  \end{tabular}
\end{table}

\begin{figure}[]
    \centering
    \includegraphics[scale=0.45]{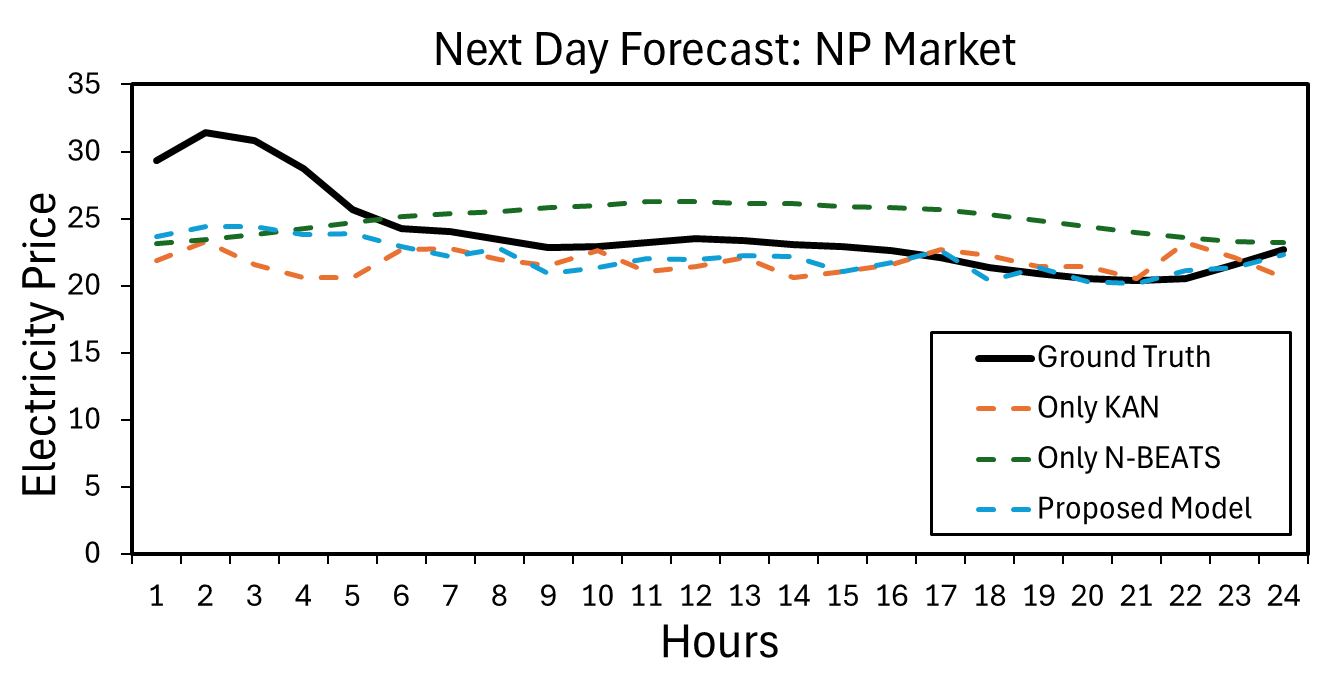}
    \caption{The next day forecast presents a comparison between the KAN, N-BEATS and Proposed model for the NP Market. As indicated, the N-BEATS model produces a smooth forecast while the proposed model uses the flexibility of the B-Spline along with the power of N-BEATS model to produce the best forecast.}
    \label{fig:NP_fcst}
\end{figure}

To present a comprehensive set of results, we conduct a series of experiments considering the Nord Pool market as the test market. The dataset is split into training and test subsets as per the method defined earlier. The hyperparameters for the proposed model and the optimization settings are optimized using a Bayesian optimization method. This method uses a tree-structured Parzen estimator to explore the hyperparameter space \cite{bergstra2013hyperopt}. All the results are reported on the NP market as the unknown or new market in a zero-shot manner. For comparison, we also repeated the experiments with standard N-BEATS architecture \cite{oreshkin2019n} and the standard KAN architecture \cite{vaca2024kolmogorov}. Table \ref{tab:zero_shot} presents a comparison between the KAN, N-BEATS and the proposed model. For each set of experiments, we consider each of the markets from the train set (FR, PJM and BE) as the primary market, and use the remaining markets from the train set as secondary. The values are averaged over different models, each time with a different market as the secondary market. For all the cases, zero-shot forecasts were generated for the NP market prices. We observe an improvement of around 13\% and 24\% in accuracy for the proposed model compared to the N-BEATS and KAN model respectively. The performance of our proposed model, in terms of forecasting errors, is within the same order of magnitude as reported in literature \cite{olivares2023neural}. Furthermore, Fig. \ref{fig:NP_fcst} shows 24-hour ahead multi step forecast generated by our proposed model, where it can be observed that the N-BEATS model tends to produce a smoother forecast while the proposed model can leverage the flexibility of the spline curve to align with the shape of the distribution.

\begin{figure}
    \centering
    \includegraphics[width=0.65\linewidth]{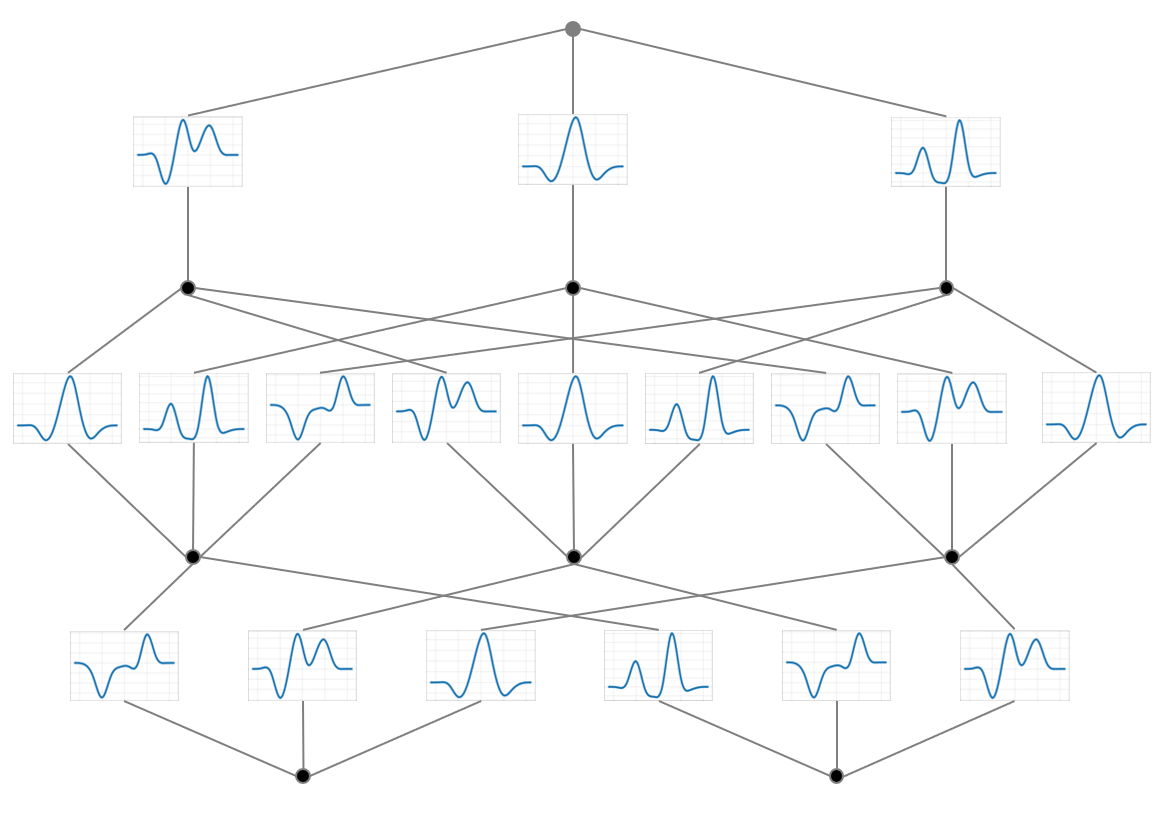}
    \caption{This representative example shows some of the functions learned by the KAN network when we do zero shot forecasting on the Nord Pool market, using France and Belgium as the primary and secondary markets repsectively. }
    \label{fig:kan_interp}
\end{figure}

In a lot of real world applications, it is important that the model used for forecasting time series data has a low inference time and is easy to scale on low resource environments. Moreover, it is important that the model is easy to understand. Since KAN uses smooth functions to approximate the time series data, KAN  based architectures make it easier for the users to interpret the forecast \cite{xu2024kolmogorov} as opposed to the large foundation models whose inner workings are difficult to understand.  Fig. \ref{fig:kan_interp} shows a sample representation of the learned functions for our proposed   multi layer KAN architecture with Nord Pool as the test market. 

\section{Conclusion}
In this paper, we present a domain adaptation framework  which uses the Kolmogorov Arnold Network with a doubly residual structure as the backbone for forecasting electricity prices for new markets. It builds on the N-BEATS like architecture with KAN at its core. Mainly composed of KAN layers the architecture is relatively light weight and fast to optimize, and has better interpretability than the MLP based models. We propose an adversarial training with two market data to generate a domain-generalized model that can then be applied to forecast prices from unseen markets in a zero-shot manner. We show the performance of the proposed method using a set of benchmark datasets from the electricity price forecasting domain and demonstrate that the proposed model outperforms the baseline model for zero shot forecasting. Although, the current model can be directly applied for several domain adaptation tasks across markets we believe that there is scope to further improve the model by incorporating external factors like weather parameters to augment the univariate features. It would be also interesting to extend the framework to allow multiple secondary markets to create a more generalized feature representation.        


\bibliographystyle{unsrt}
{\small

}

\newpage
\appendix
\section{Proposed Framework}
The objective of our proposed framework is to learn a generalized representation of the price data based on the primary and secondary markets while preserving a low risk on the primary (supervised) market. In this section, we describe the underlying concepts and the framework. 

\subsection{Domain Generalization}
We assume that we have two established markets with significant historical data. We plan to tackle the challenge of learning generalized representation by using the supervised error on the primary market and a notion of the distance between the source and target distributions. This enables the model to learn \emph{Domain Invariant features}. This term is usually defined as features which if given to a classifier, the classifier should not be able to predict which domain or market the features originate from \cite{ben2006analysis}. In addition to the domain invariant features, we use the supervised training regime to learn \emph{Domain Specific features}. These features are specific to the primary market and are discriminative in nature. Unlike previous work which worked with fixed feature representations, the proposed model learns both domain invariant and domain specific features within the same end-to-end training process. 

Let $\mathbb{X} :=\mathbb{R}_{\alpha}$ and $\mathbb{Y} := \mathbb{R}_{\beta}$ be the price series data and the output values respectively. The look back period and horizon are denoted by $\alpha$ and $\beta$ respectively. In the problem formulation we consider 2 energy markets, primary $\mathbb{M}_p$ and  secondary market $\mathbb{M}_s$. The new market which has limited or no data is denoted by $\mathbb{M}_t$. Let's define the set of of all Borel joint probability measures on $\mathbb{X} \times \mathbb{Y}$ by $P:= P (\mathbb{X} \times  \mathbb{Y})$. Then the marginal probabilities on $\mathbb{X}$ and $\mathbb{Y}$ are denoted by $\mathbb{P}_{\mathbb{X}}$ and $\mathbb{P}_{\mathbb{Y}}$ respectively. On the same lines, the marginal probability of $\mathbb{M}_{p}$ is denoted by $\mathbb{P}_{p}$  and for $\mathbb{M}_s$ by $\mathbb{P}_s$ where $\mathbb{P} \in \textit{P}$. 

During the training process, our framework jointly optimises a time series forecasting model and a discriminative classifier. The purpose of the forecasting model is to forecast the energy prices for the input market. While the purpose of the classifier is to identify the market of origin from the input features. To be aligned with literature, we denote the classifier as the \emph{Domain Classifier}. On one hand, the parameters of the forecasting model are optimized in order to minimize the error on the primary energy market. On the other hand, the parameters of the generalized mapping are optimized to reduce the loss on the primary market and maximize the loss of the domain classifier.
Let the loss function be defined as $\mathcal{L} : \mathbb{Y} \times \mathbb{Y} \rightarrow \mathbb{R}$. The objective of the proposed framework is to build a forecasting model $\mathcal{F} : \mathbb{X} \rightarrow \mathbb{Y}$  such that the risk on the new market is minimized with no information about the forecast of $\mathbb{M}_T$
\begin{equation}
\centering
\mathbb{R}_{\mathbb{M}_{T}} (\mathcal{F}) = \mathbb{P}(\mathcal{F}(X) \ne Y)
\end{equation}

\begin{figure*}[!h]
\centering
    \includegraphics[width=0.8\linewidth]{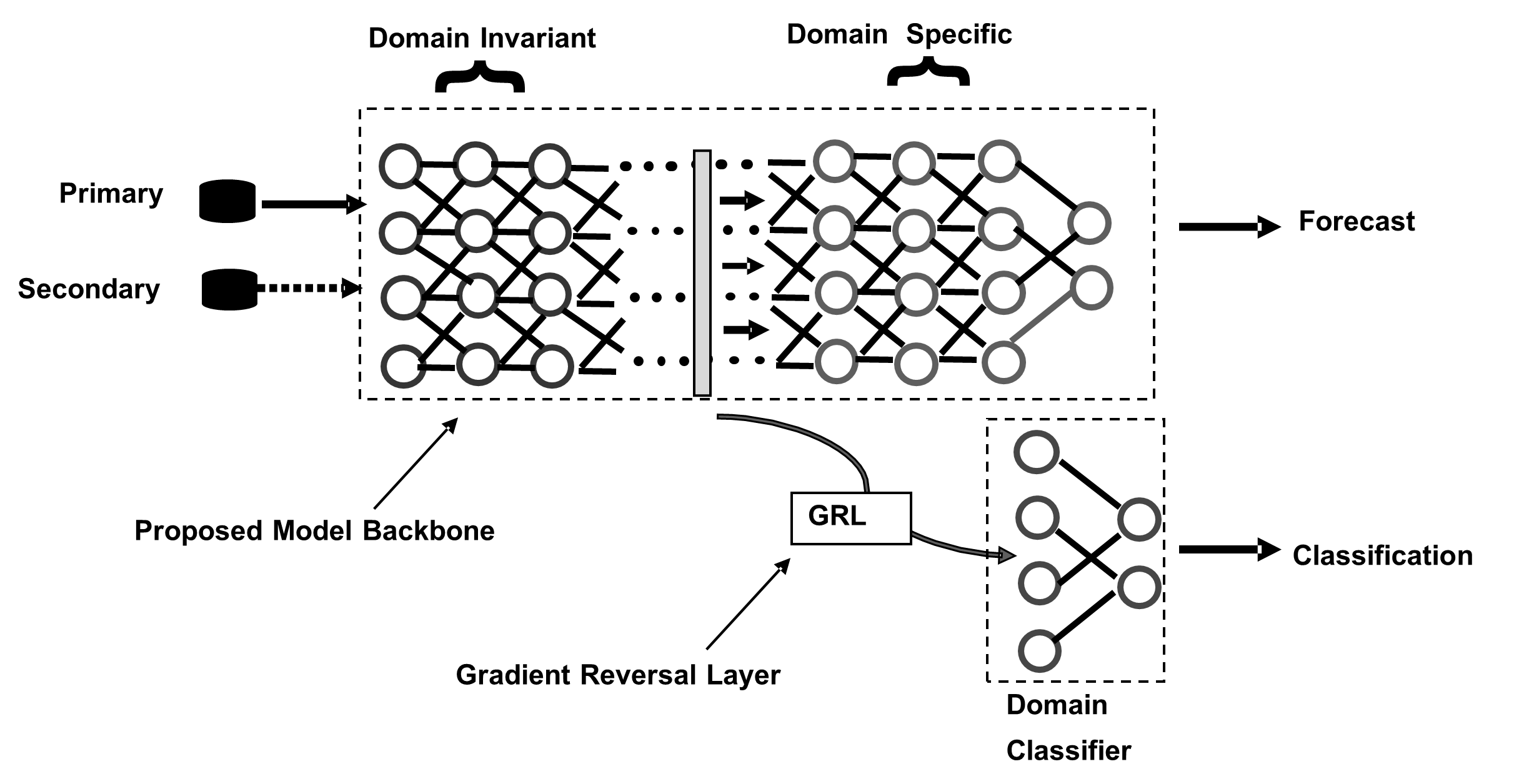}
    \caption{Schematic of our Domain Adaptation Framework.}
    \label{fig:framework_schematic}
\end{figure*}

This approach of jointly minimizing one loss while maximizing another loss is denoted in literature as \emph{Adversarial training}. The framework is described at a high level in this schematic \ref{fig:framework_schematic}.

The adversarial nature of the training enables the model to learn a generalized representation. Most importantly, the proposed framework performs all the functions in a single training process in an end-to-end manner. 
To implement the adversarial training between the forecasting model and the domain classifier, we use a \emph{gradient reversal layer}. This approach was first proposed by Ganin \textit{et.al.} \cite{ganin2016domain}. The gradient reversal layer is designed in a manner such that it does not change the input during forward propagation and reverses the gradient by multiplying it by a negative scalar during propagation. The proposed framework can be used as a generic approach for electricity price forecasting of new markets because it uses standard loss functions and uses widely used Adam gradient descent algorithm for training. 

\subsection{Proposed Model Backbone}
The proposed model backbone explores the application of Kolmogorov Arnold networks with a doubly residual architecture inspired by the NBeats architecture \cite{oreshkin2019n}. It consists of a stacked KAN network with learnable activation function at the edges. The learnable activation functions are trained using back-propagation. The network decomposes the time series by creating local nonlinear projections of the data onto B Spline functions across different blocks. Each block learns expansion coefficients for the backcast and forecast elements. For $M,L \in \mathbb{N}$, the model comprises M stacks with each stack consisting of L blocks. The blocks share the same activation functions within each respective stack and are recurrently operated based on the doubly residual stacking principle. 
The output of each forecast $\widehat{y}$ is obtained through hierarchical aggregation of each block's forecast and the last backcast derived by a residual sequence of blocks serves as an input to the next stack.
For a given stack $i$ and a block $j$ within it, the model takes the input data and passes it through the KAN network to learn the parameters of the B Spline for the forecast and backcast coefficients denoted by $\theta^{f}_{i,j}$ and $\theta^{b}_{i,j}$ respectively. Let the dimension of the hidden unit be $N_h$ and the stack basis be $N_s$. Then we know $\theta^{f}_{i,j} \in \mathbb{R}^{N_s}$ , $\theta^{b}_{i,j} \in \mathbb{R}^{N_s}$  and $h_{i,j} \in \mathbb{R}^{N_h}$. 
\begin{equation}
\centering
h_{i,j} = NN_{i,j} (y^{b}_{i,j-1},X_{b-1})
\end{equation}

\begin{equation}
\begin{aligned}
\theta^{f}_{i,j} = NN_{linear}^{f}(h_{i,j})\\
\theta^{b}_{i,j} = NN_{linear}^{b}(h_{i,j})
\end{aligned}
\end{equation}

These learned coefficients are used to generate the backcast $\hat{y}^{b}_{i,j}$ and forecast  $\hat{y}^{f}_{i,j}$ components.
\begin{equation}
\begin{aligned}
\hat{y}^{f}_{i,j} = \mathbb{H}^{f}_{i,j} \theta^{f}_{i,j}\\
\hat{y}^{b}_{i,j} = \mathbb{H}^{b}_{i,j} \theta^{b}_{i,j}
\end{aligned}
\end{equation}

where $\mathbb{H}^{f}_{i,j} \in \mathbb{R}^{LX N_{s}}$ and $\mathbb{H}^{b}_{i,j} \in \mathbb{R}^{L X N_{h}}$ denote the basis vectors of the blocks.

The doubly residual principle is used between stacks. The price series is sequentially decomposed by  subtracting the predicted backcast $\hat{y}^{b}_{i,j}$ from the original series to obtain the next series $y^{b}_{i,j+1}$. The forecast for the stack is the aggregate of the block forecasts.
\begin{equation}
\centering
y^{b}_{i,j+1} = y^{b}_{i,j} - \hat{y}^{b}_{i,j}
\centering
\hat{y}^{f} = \sum^{B}_{j=1}{\hat{y}^{f}_{i,j}}
\end{equation}
The final forecast is the hierarchical aggregation of all the stacks after the residual operations are completed. 
\begin{equation}
\centering
\hat{y}^{f} = \sum^{S}_{i=1}{\hat{y}^{f}_{i}}
\end{equation}
where $B$ and $S$ denote the number of blocks and stacks respectively.

\section{Zero Shot Learning}
We begin our analysis of Zero Shot Learning with attributes by noting that it is a two-stage process comprised of a training phase and an inference phase. In the training phase, we use the seen classes to learn a map from the items to the attribute space, and in the inference phase, we use the class-attribute matrix to infer the correct class given the item-attribute representation. We can identify two types of errors using this two-stage decomposition. The first type is caused by domain shift and is related to the training phase. The map from items to attribute space that has been trained on seen classes may not generalize well to unseen classes. 

\subsection{Covariate Shift}
In covariate shift, we assume that the marginal distribution of source and target data change while the predictive dependency remains unchanged. One possible way to solve covariate shift is reweighting scheme. 

\begin{equation}
\begin{aligned}
    R_T^l(h) = \mathbf{E}_{(x,y)\sim T} l(h(\textbf{x},y)\\
             = \mathbf{E}_{(x,y)\sim T} \frac{S(x,y)}{S(x,y)}l(h(\textbf{x},y)\\
             =\sum_{(x,y)\in XxY} \mathbb{T}(x,y) \frac{S(x,y)}{S(x,y)} l(h(\textbf{x},y)\\
             =\mathbf{E}_{(x,y)\sim S}\frac{\mathbb{T}(x,y)}{S(x,y)} l(h(\textbf{x},y) 
\end{aligned}
\end{equation}
Using the assumption $S(y|x) = T(y|x)$, we get
\begin{equation}
\begin{aligned}
    R_T = \mathbb{E}_{(x,y)\sim S}\frac{T_X(x)T(y|x)}{S_X(x)S(y|x)} l(h(\textbf{x},y) \\ 
    =\mathbb{E}_{(x,y)\sim S}\frac{T_X(x)}{S_X(x)}l(h(\textbf{x},y)
\end{aligned}
\end{equation}

\section{Evaluation Metrics}
We use widely accepted practice of evaluating the accuracy of point forecast with the mean absolute error (MAE), mean and symmetric mean absolute error (sMAPE). MAE is used to compare the absolute forecasting errors between markets. However, it assumes that the electricity prices across the markets are in a comparable range. While the symmetric mean absolute percentage error (SMAPE) solves this problem, it has a distribution with undefined mean and infinite variance \cite{hyndman2006another}.

\begin{equation}
MAE = \displaystyle\frac{1}{n}\sum_{i=1}^{n}|y_i - \hat{y}_i|
\end{equation}

\begin{equation}
\text{SMAPE}= \frac{1}{n} \sum_{i=1}^{n} \frac{ 2*|y_i - \hat{y}_i|}{|y_i| + |\hat{y_i}|}
\end{equation}

\end{document}